\documentclass[
]{ceurart}

\usepackage{listings}
\usepackage{subcaption}

\begin{document}

\copyrightyear{2026}
\copyrightclause{Copyright for this paper by its authors.
  Use permitted under Creative Commons License Attribution 4.0
  International (CC BY 4.0).}

\conference{1st Streaming Continual Learning Bridge at AAAI26, January 21, 2026, Singapore.}

\title{Tracking Adaptation Time: Metrics for Temporal Distribution Shift}
\author[1]{Lorenzo Iovine}[%
orcid=0009-0007-8974-8638,
email=lorenzo.iovine@polimi.it,
]
\cormark[1]
\address[1]{Politecnico di Milano, DEIB, Via Giuseppe Ponzio, 34, 20133 Milan, Italy }

\author[1]{Giacomo Ziffer}[%
orcid=0000-0002-2768-3580,
email=giacomo.ziffer@polimi.it,
]

\author[1]{Emanuele {Della Valle}}[%
orcid=0000-0002-5176-5885,
email=emanuele.dellavalle@polimi.it,
]

\cortext[1]{Corresponding author.}

\begin{abstract}
Evaluating robustness under temporal distribution shift remains an open challenge. Existing metrics quantify the average decline in performance, but fail to capture how models adapt to evolving data. As a result, temporal degradation is often misinterpreted: when accuracy declines, it is unclear whether the model is failing to adapt or whether the data itself has become inherently more challenging to learn. In this work, we propose three complementary metrics to distinguish adaptation from intrinsic difficulty in the data. Together, these metrics provide a dynamic and interpretable view of model behavior under temporal distribution shift. Results show that our metrics uncover adaptation patterns hidden by existing analysis, offering a richer understanding of temporal robustness in evolving environments.
\end{abstract}


\maketitle

\section{Introduction}

Machine Learning (ML) models are typically evaluated under the assumption that the training and test data are drawn from the same underlying distribution. However, in real-world applications, this assumption rarely holds. Data distributions often evolve over time due to changes in the environment, user behavior, or underlying data-generating processes, a phenomenon known as \textbf{Temporal Distribution Shift}. Such shifts challenge the static nature of conventional learning approaches, which are usually designed for stationary settings. 

This problem has recently gained attention through benchmarks such as Wild-Time~\cite{yao2022}, which exposes the inability of modern models to maintain consistent performance when evaluated on future data. These benchmarks typically report two key metrics: \textbf{In-Distribution (ID) accuracy}, measured on samples from the same time period as training, and \textbf{Out-Of-Distribution (OOD) accuracy}, computed as the average performance over subsequent time periods. Empirically, these studies reveal a persistent gap between ID and OOD accuracy, suggesting that existing models struggle to generalize or adapt over time.

While this gap is widely interpreted as evidence of poor temporal adaptation, we believe that this interpretation is ambiguous. The ID-OOD gap combines two fundamentally different factors:
\begin{itemize}
    \item \textbf{Adaptation lag - }when the model fails to incorporate temporal changes, resulting in a delayed response to distribution shifts \emph{(well recognized)}. 
    \item \textbf{Intrinsic difficulty in the data - }when later time periods contain less separable or noisier samples, both ID and OOD performance degrade, even if the model adapts correctly \emph{(largely underestimated)}.
\end{itemize}

The \textbf{problem} we address in this work is precisely this ambiguity: existing evaluation metrics are unable to distinguish whether performance degradation over time is caused by increasing data difficulty or by insufficient model adaptation.
This ambiguity limits our understanding of model behavior under temporal shift and hinders the design of adaptive learning algorithms.

\begin{figure*}[t]
    \centering
    \includegraphics[width=0.33\textwidth]{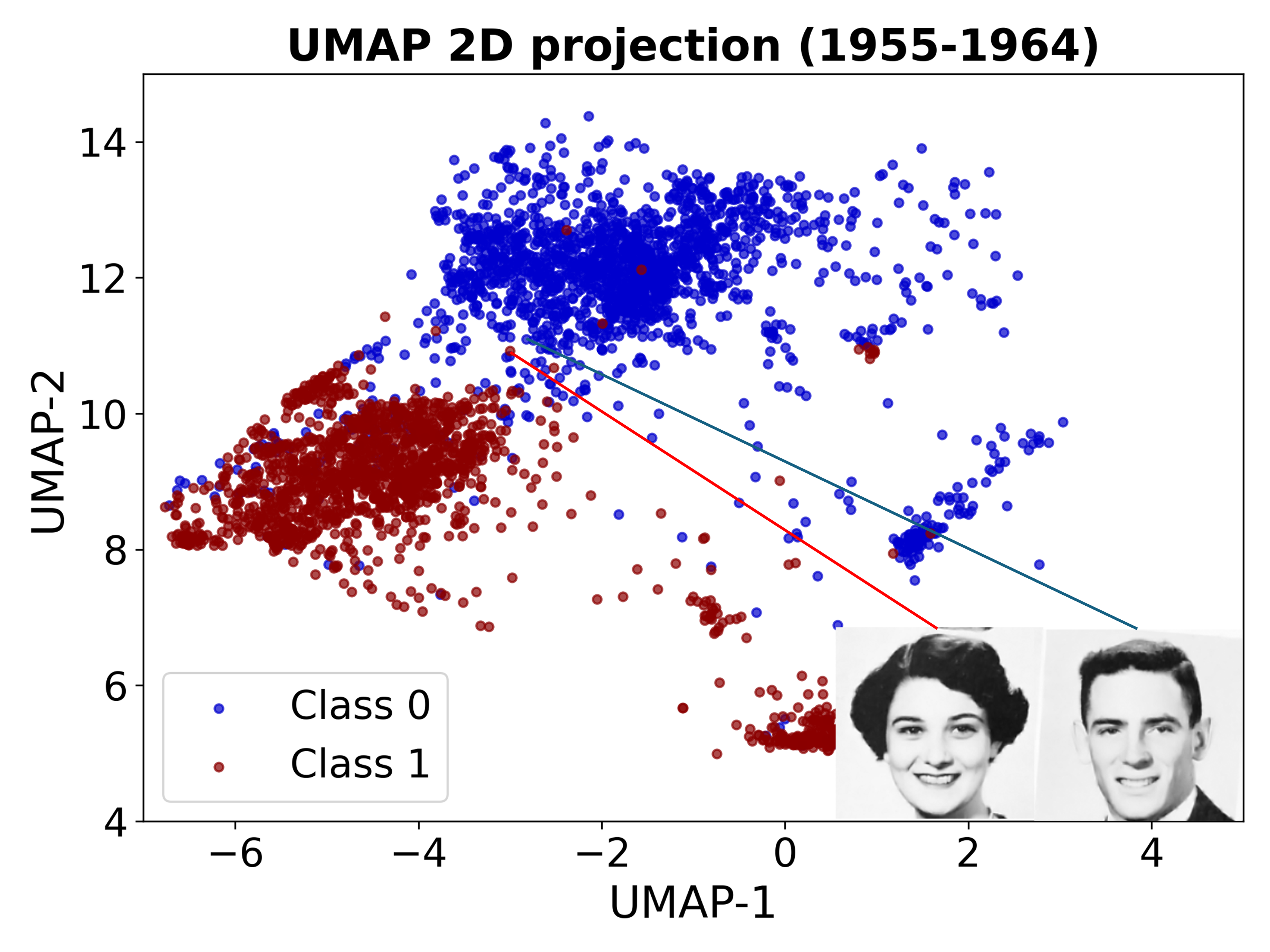}
    \includegraphics[width=0.33\textwidth]{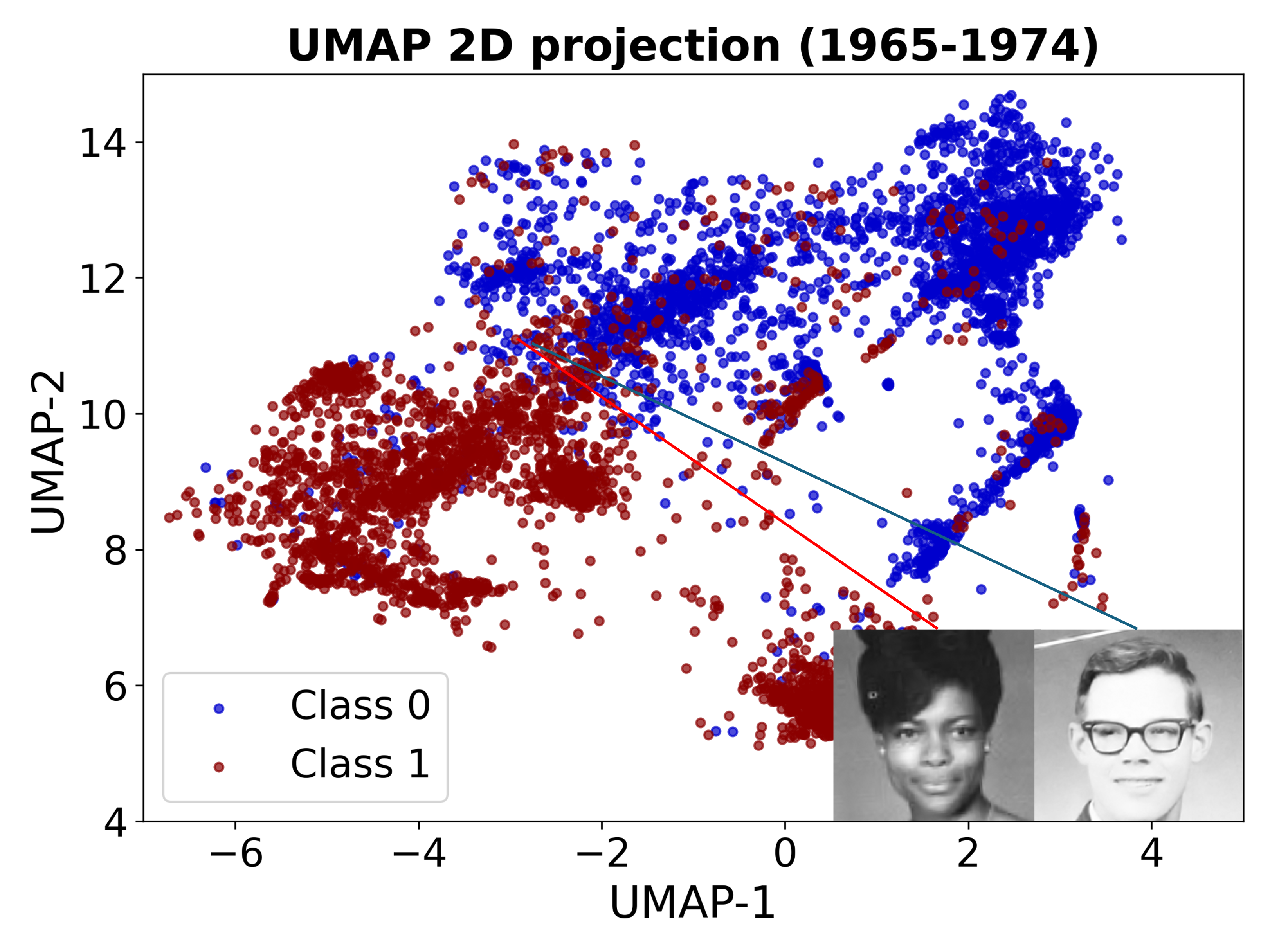}
    \includegraphics[width=0.33\textwidth]{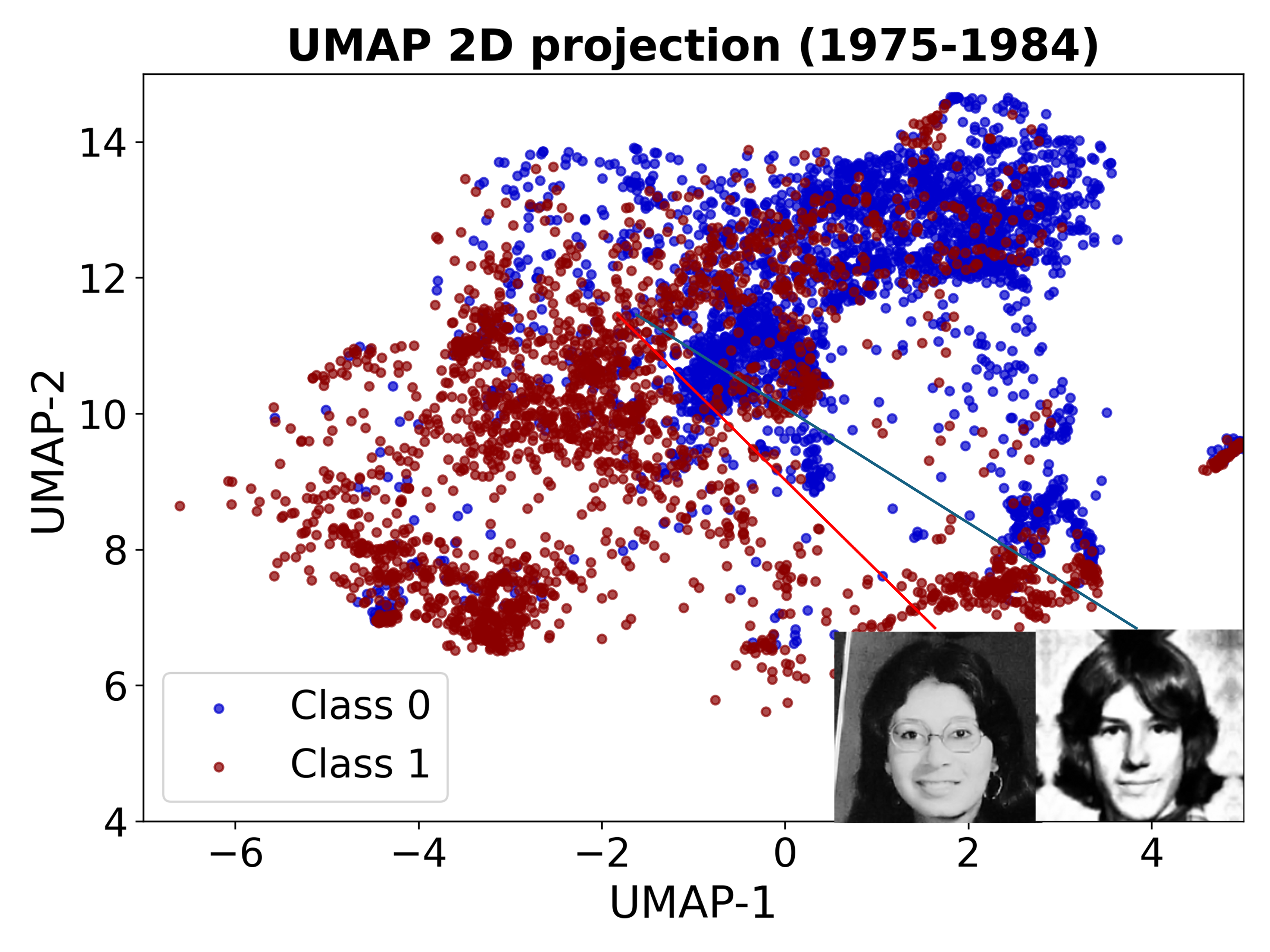}
    \caption{
    UMAP 2D projections of Yearbook feature embeddings across consecutive decades.
    Blue and red points correspond to different classes (e.g., male vs. female).
    Over time, the class clusters — initially well separated — become increasingly overlapping,
    reflecting higher intrinsic data difficulty and complex temporal changes.
    This visual evidence supports our hypothesis that performance degradation is not solely due to limited model generalization, but also to evolving data complexity.
    }
    \label{fig:umap_yearbook}
\end{figure*}

To illustrate this phenomenon, Figure~\ref{fig:umap_yearbook} shows a two-dimensional UMAP~\cite{mcinnes2018} projection of feature representations extracted from the Yearbook dataset~\cite{ginosar2015}, grouped in 10-year intervals.
Initially (1955–1964), classes remain largely separable, with clear boundaries between male and female portraits.
However, as time progresses, the clusters gradually overlap, indicating that visual features from different years become increasingly intertwined, a possible consequence of emerging visual diversity, with previously uncommon facial traits and aesthetics becoming more frequent.
This suggests that the drop in model performance observed in later years does not necessarily arise from poor generalization, but from a genuine increase in data difficulty, the classes themselves become less separable as the underlying generative process evolves.
In other words, temporal degradation often reflects complex relations between data and the problem, rather than a failure to adapt.

In this work, we \textbf{contribute by introducing} three complementary metrics to address this limitation. Together, these metrics isolate the model's temporal adaptation dynamics from the intrinsic evolution of data difficulty. We demonstrate the effect of our metrics on datasets from Wild-Time. Our results suggest that what appears as a persistent ID-OOD gap in prior work often reflects a temporal lag in adaptation rather than a complete failure to generalize, motivating a reevaluation of how we measure and compare models under temporal distribution shift. To facilitate adoption and reproducibility, we release the code to compute the proposed metrics.\footnote{\url{https://github.com/lorenzoiovine99/Metrics-for-Temporal-Distribution-Shift}}

\section{Related Work}
\textbf{Temporal distribution shift} refers to changes in the data distribution that occur as time progresses.
Unlike domain shifts, where different datasets represent distinct environments or acquisition sources, temporal shifts emerge naturally as the world evolves: sensor properties change, user behavior drifts, visual or textual styles shift, and contextual conditions evolve. Large-scale benchmarks such as WILDS~\cite{koh2021} and Wild-Time have made this setting explicit by organizing data along temporal axes. Datasets like FMoW-Time (a subset of the Functional Map of the World dataset~\cite{ChristieFWM18}) and Yearbook expose temporal evolution through gradual changes in content and context, making them ideal testbeds for studying model robustness to time-dependent drift.
These works have shown that even high-capacity models exhibit substantial degradation when evaluated on future samples, underscoring the challenge of temporal generalization. However, most temporal benchmarks measure only static performance snapshots at different time points.
Such evaluations quantify accuracy decay but fail to reveal how performance changes unfold, whether due to intrinsic task difficulty or to a lack of temporal adaptation.
This limitation motivates the need for metrics that explicitly capture the temporal dynamics of adaptation rather than static robustness summaries.

The broader concept of \textbf{concept drift} originates from the data stream mining literature~\cite{GamaZBPB2014,ZliobaitePG2015}.
It denotes any change in the joint distribution P(X,Y) that affects a model’s predictive behavior over time or across contexts.
Such changes can stem from variations in the data-generating process, user populations, environmental conditions, or even shifts in the underlying concept being learned.
In other words, while temporal distribution shift specifically refers to distributional changes induced by time, concept drift can occur for many other reasons — temporal, environmental, behavioral, or stochastic. Research on concept drift has developed adaptive methods capable of detecting and responding to such changes online. Classical algorithms, like ADWIN~\cite{bifet2007}, monitor error rates or distributional statistics to detect significant drift, while streaming and Continual Learning approaches such as EWC~\cite{Kirkpatrick2017}, SI~\cite{zenke2017}, and A-GEM~\cite{chaudhry2018} introduce mechanisms to balance plasticity and stability in non-stationary environments.
Despite this, most evaluations in these settings focus on aggregate measures, such as average accuracy, forgetting, or backward transfer, that summarize long-term performance rather than capturing fine-grained temporal adaptation behavior.
Consequently, even in the context of streaming and continual learning, the community lacks metrics that quantify how effectively a model adapts to evolving data distributions over time.

\textbf{Evaluating models} under non-stationarity has traditionally relied on metrics designed for discrete tasks or static shifts.
In Continual Learning~\cite{YenYZ2018} forward and backward transfer measures were introduced to quantify interference and knowledge retention across tasks~\cite{lopez2017}.
While effective for task-structured evaluation, these metrics are ill-suited for continuous temporal evolution, where boundaries between distributions are not well-defined. Temporal benchmarks, on the other hand, often report aggregated accuracy or ID–OOD comparisons across years. Although informative, these metrics cannot distinguish between degradation due to intrinsic difficulty (e.g., increased noise or reduced separability) and degradation caused by insufficient adaptation.
Both scenarios can produce similar accuracy curves but reflect fundamentally different model behaviors. Similar limitations have been noted in domain generalization and robustness research~\cite{taori2020}, where performance gaps may conflate distribution hardness with model capacity.
In temporal settings, this ambiguity obscures our understanding of whether a model fails because the data becomes harder, or because it cannot adjust quickly enough to a changing environment. Hence, there remains a need for evaluation metrics that explicitly capture temporal adaptation dynamics, describing how rapidly, stably, and persistently a model adapts as data distributions evolve.
Addressing this gap is the central focus of this work.

\section{Problem Statement}
We consider a supervised learning setting where data distributions evolve over time. Let \begin{equation}D_t=\{(x_i,y_i)\}_{i=1}^N,\end{equation} denote the dataset available at time t, sampled from an underlying joint distribution $P_t(X,Y)$.
In the presence of temporal distribution shift, these distributions can change as t increases.
This setting reflects many real-world scenarios in which models are trained on past data and then deployed in the future: the world changes, and the model gradually becomes misaligned with the data it encounters.

Existing benchmarks, such as Wild-Time, summarize this phenomenon through In-Distribution (ID) and Out-of-Distribution (OOD) accuracy: performance on data from the training period versus future periods.
While intuitive, this pair of metrics provides only a static picture of what is inherently a dynamic process.
A gap between ID and OOD accuracy can be observed even when the model adapts correctly, simply because future data is intrinsically harder (less separable, noisier, or more ambiguous).
Conversely, two models with identical ID–OOD gaps might exhibit entirely different temporal behaviors: one may degrade immediately and never recover, while another might adapt rapidly to new conditions.
Therefore, these static measures conflate intrinsic difficulty with adaptation capability, making it impossible to determine why performance changes through time.

What is missing is a principled way to evaluate how models adapt, not only how much their performance drops.
We argue that measuring robustness under temporal shift should involve characterizing the temporal adaptation dynamics.
Understanding these dynamics is essential for comparing algorithms designed for continual, adaptive, or streaming learning, and for identifying the true limitations of current approaches.

This leads to our central \textbf{research question}:
\textit{How can we quantitatively distinguish between degradation caused by intrinsic data difficulty and degradation caused by insufficient temporal adaptation in machine learning models?}
Addressing this question requires moving beyond static ID/OOD metrics toward a framework that explicitly captures adaptation over time.

\section{Proposed Metrics}
To characterize how models evolve over time, we introduce three complementary post-hoc metrics that quantify distinct aspects of temporal adaptation. They capture how long a model remains valid after training, when degradation becomes evident, and how effectively it adapts to future data. All metrics are computed after the full temporal sequence has been observed, allowing a retrospective evaluation of how well a model trained at time t would have matched the performance of an "oracle" model retrained on the data distribution of each target time $\tau$. Formally, the oracle represents the maximum achievable performance for a given period under ideal adaptation. By normalizing a model's temporal performance against this reference, our metrics isolate the degradation due to lack of adaptation from that caused by increasing data difficulty.

\subsection{Temporal Transfer Ratio (TTR)}
Let $A(t,\tau)$ denote the accuracy obtained when a model is trained on data from time t and evaluated on data from time $\tau$. We define the \textit{Temporal Transfer Ratio (TTR)} as: \begin{equation}
    g(t,\tau)=\frac{A(t,\tau)}{A(\tau,\tau)}.
\end{equation}
The denominator $A(\tau,\tau)$ represents the "oracle" performance level, the accuracy that would be achievable if the model were trained directly on the target time.
To ensure interpretability, we clip $g(t,\tau)$ to 1 in cases where a model trained at time t outperforms the oracle trained at $\tau$, enforcing $g(t,\tau)\in[0,1]$.
It quantifies how much of the realistic maximum at time $\tau$ is preserved by a model trained in the past. A value close to 1 indicates strong temporal transfer, while lower values indicate growing misalignment between the model and the evolving data distribution. This function serves as the foundation for all three proposed metrics.

\subsection{Stability Horizon (SH)}
The \textit{Stability Horizon (SH)} captures how long a model trained at time t remains reliable before its performance drops below an acceptable level. Formally, for a chosen tolerance threshold $\delta\in[0,1]$:
\begin{equation}
    SH_\delta(t) = max\{h\ge0:g(t,t+h)\ge\delta\}.
\end{equation}
Intuitively, $SH_\delta(t)$ measures the number of future time steps for which a model trained at t maintains at least a fraction $\delta$ of the oracle accuracy. This provides a direct estimate of the temporal validity window of the model, the time span during which it can be deployed without retraining.
The threshold $\delta$ can be adapted to the application: for instance, $\delta=0.9$ corresponds to a 10\% acceptable loss in accuracy.

\subsection{Drift Horizon (DH)}
While the Stability Horizon identifies when performance falls below a target level, the \textit{Drift Horizon (DH)} detects when that drop becomes statistically significant.
We define a cumulative drift statistic as:
\begin{equation}
    \begin{gathered}
    S_0 = 0, \\
    S_h = \max(0, S_{h-1} + (|A(t,t+h) - A(t,t)| - \epsilon)).
    \end{gathered}
\end{equation}
where $\epsilon$ is a small tolerance parameter that filters out random fluctuations. The DH is then the smallest temporal distance h for which the cumulative deviation exceeds a significance threshold $\lambda$:
\begin{equation}
    DH_t=min\{h\in[1,H]|S_h > \lambda\},
\end{equation}
where H is the maximum evaluation horizon (number of future steps available).
Intuitively, $DH_t$ answers the question: \textit{if I stop updating the model after a given time step, after how long will performance degradation become statistically evident?}
This metric captures the onset of observable drift and provides a temporal notion of performance stability that is less dependent on arbitrary accuracy thresholds.

\subsection{Temporal Adaptation Score (TAS)}
The \textit{Temporal Adaptation Score (TAS)} measures how effectively a model trained at time t generalizes to future periods relative to their intrinsic difficulty. For each training time t, we compute the average OOD accuracy over the next n time steps and normalize it by the average oracle accuracy on those same time steps:
\begin{equation}
    \begin{gathered}
    \overline{A}_{OOD}(t)=\frac{1}{n}\sum_{k=1}^n A(t,t+k), \\
    \overline{A}_{ID}(t)=\frac{1}{n}\sum_{k=1}^n A(t+k,t+k), \\
    TAS_t=\frac{\overline{A}_{OOD}(t)}{\overline{A}_{ID}(t)}.
    \end{gathered}
\end{equation}

As TAS is derived from averaged TTR values, it inherits the same clipping strategy. A TAS close to 1 indicates that the model achieves nearly the same performance as an oracle retrained for all future time steps, suggesting strong adaptation across time. Lower TAS values signal limited adaptability or growing temporal misalignment. Unlike the ID-OOD gap, TAS captures relative adaptation: how well a model follows the evolving oracle rather than its absolute performance drop.

\subsection{Interpretation and Comparative Insights}
Each proposed metric provides a complementary view of temporal adaptation:
\begin{itemize}
    \item \textbf{TAS} measures how well a model follows the oracle, i.e., its \textbf{relative adaptation to future data}.
    \item \textbf{SH} measures for how long the model remains above an acceptable performance threshold, i.e., its \textbf{temporal validity}.
    \item \textbf{DH} measures after how many time steps performance degradation becomes statistically evident, i.e., its \textbf{sensitivity to drift}.
\end{itemize}

When comparing models, these metrics enable nuanced interpretations. Two models with similar average OOD accuracy may exhibit very different TAS values, revealing which model better maintains relative performance to the oracle across time. Conversely, models with similarly TAS may differ in SH or DH, distinguishing those that remain stable for longer periods from those that degrade more abruptly. Together, TAS, SH, and DH form a coherent metric suite that separates adaptation from difficulty, providing interpretable temporal diagnostics that static metrics such as ID-OOD accuracy cannot capture.

\section{Experimental Setup and Results}
We empirically evaluate the proposed metrics to assess their ability to disentangle temporal adaptation from intrinsic data difficulty. Our evaluation is designed to answer two main questions: (i) whether the metrics provide additional insight beyond standard ID–OOD accuracy comparisons, and (ii) whether they consistently characterize temporal adaptation dynamics across datasets with different types of temporal shifts.

To this end, we conduct experiments on two benchmarks from the Wild-Time suite, Yearbook and FMoW-Time, which exhibit complementary temporal behaviors. We compare multiple learning paradigms and analyze the proposed metrics both quantitatively and qualitatively, highlighting how they capture stability, drift onset, and relative adaptation over time.

\subsection{Datasets}
We evaluate the proposed metrics on two temporal benchmarks from the Wild-Time suite: \textbf{Yearbook} and Functional Map of the World-Time (\textbf{FMoW-Time}).
Both datasets are explicitly organized along a temporal axis, allowing a controlled analysis of model behavior under real temporal distribution shifts.
Yearbook consists of grayscale portraits of American high school students from 1930 to 2013. Each year defines a distinct data subset, and the task is gender classification. Temporal shifts reflect gradual changes in photographic style, lighting, and fashion trends rather than abrupt domain changes.
FMoW-Time contains satellite images of land-use scenes captured over multiple years and geographic regions. The task is to classify the functional category of each scene (e.g., airport, hospital, residential area). Temporal shifts arise from environmental changes, sensor updates, and evolving land use, producing a rich and realistic testbed for long-term adaptation analysis.

While Yearbook exhibits smooth, visually interpretable drifts dominated by the evolution of P(X), FMoW features complex multimodal shifts combining temporal and spatial factors. Together, they provide complementary perspectives on temporal robustness: gradual aesthetic drift versus heterogeneous real-world dynamics.

\subsection{Models and Evaluated Metrics}
We evaluate several models representative of different learning paradigms. Empirical Risk Minimization (ERM), the baseline model trained independently on each year's data. CORAL, a domain alignment method minimizing feature-level covariance shift. Elastic Weight Consolidation (EWC), a continual learning regularization approach that penalizes the deviation from previous parameters. Synaptic Intelligence (SI), an alternative regularization-based continual learner. Fine-Tuning (FT), sequentially updates the model using new data without explicit drift control. And, finally, a pipeline which combines Momentum Contrastive Learning techniques~\cite{he2020} with Streaming Machine Learning~\cite{GomesRBBG19} models, designed to improve temporal adaptability~\cite{iovine2025}.

For FMoW, we computed ID and OOD accuracy, TAS, SH and DH. For Yearbook we focused on ID, OOD and TAS. All experiments were re-run from scratch using the official Wild-Time implementation. The combination of the metrics enables both global and dynamic analysis: ID-OOD scores reveal absolute robustness, whereas TAS-SH-DH expose temporal adaptation dynamics.

\subsection{Results on FMoW-Time}

\begin{figure*}[t]
    \centering
    \begin{subfigure}{0.5\textwidth}
        \centering
        \includegraphics[width=\linewidth]{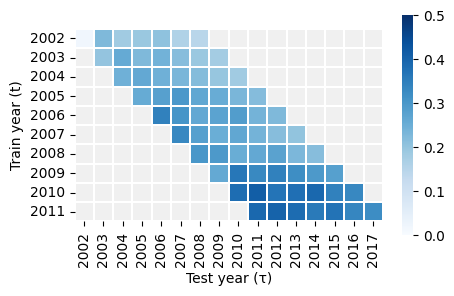}
        \caption{\(A(t,\tau)\): Accuracy}
        \label{fig:matrices-acc}
    \end{subfigure}\hfill
    \begin{subfigure}{0.5\textwidth}
        \centering
        \includegraphics[width=\linewidth]{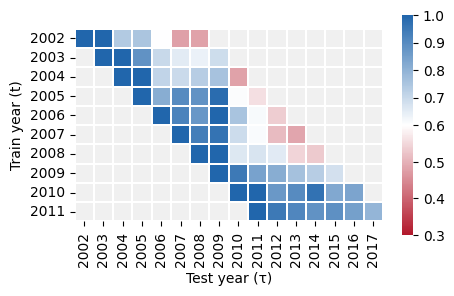}
        \caption{\(g(t,\tau)\): TTR}
        \label{fig:matrices-g}
    \end{subfigure}
    \caption{
        Temporal performance of the Fine-Tuning model on FMoW. 
        Each row corresponds to a model trained at year \(t\) and evaluated on future years \(\tau>t\).
        (a) \(A(t,\tau)\): darker cells indicate \emph{higher accuracy}; grey cells mark non-evaluated pairs. 
        (b) \(g(t,\tau)\): the colormap is centered at the tolerance threshold \(\delta=0.6\); \emph{white} denotes \(g(t,\tau)=0.6\) exactly, \emph{blue} values \(>\!0.6\) indicate smaller deviation from the oracle (\(g\!\approx\!1\)), and \emph{red} values \(<\!0.6\) indicate larger deviation. Crossing below \(\delta\) marks the end of the Stability Horizon for the model trained at \(t\).
    }
    \label{fig:matrices}
\end{figure*}

To illustrate how the proposed metrics are derived, Figure~\ref{fig:matrices} shows the temporal accuracy matrices \(A(t,\tau)\) and their normalized counterparts \(g(t,\tau)\) for the \textit{Fine-Tuning (FT)} model on FMoW-Time. 
Each row corresponds to a model trained at time \(t\) and evaluated across future years \(\tau > t\).
The normalized matrix \(g(t,\tau)\) captures how well the model maintains its relative performance over time, with darker cells indicating a smaller deviation from the oracle.

In our experiments with this dataset, we configured the parameters of the proposed metrics to reflect realistic temporal dynamics in the dataset. Specifically, we set the tolerance threshold for the Stability Horizon to $\delta=0.6$, and the deviation threshold for the Drift Horizon to $\lambda=0.15$.
These values were chosen to balance sensitivity and robustness, ensuring that minor fluctuations in yearly performance do not dominate the evaluation while still capturing substantial temporal drifts.

For Fine-Tuning, the computed Stability and Drift Horizons are reported per year below, showing how long the model remains reliable after training and when performance degradation becomes evident:
\[
SH_t = [4, 6, 5, 5, 5, 4, 4, 6, 6, 6] \quad (\text{average } 5.1~\text{years})
\]
\[
DH_t = [2, 7, 7, 7, 6, 7, 7, 7, 7, 7] \quad (\text{average } 6.4~\text{years})
\]
We use 7 as a sentinel value that indicates that the threshold was not crossed within the observable window (H=6).
In these cases, the value is truncated to the maximum observable window, meaning that the model maintained acceptable performance or did not show statistically significant drift throughout the evaluation period.
This behavior highlights temporal persistence rather than an absolute duration beyond seven years.
Overall, this pattern indicates that FT maintains acceptable performance for approximately five years on average, while significant drift is only detected after about six years, suggesting relatively stable behavior over time.

\begin{table}[b]
  \caption{Comparison of models on the FMoW dataset. 
While ID and OOD accuracies reflect overall performance, TAS, SH, and DH reveal complementary aspects of temporal adaptation and stability.}
  \label{table1}
  \begin{tabular}{l|l|l|l|l|l}
    \toprule
    \textbf{Model} & \textbf{ID} & \textbf{OOD} & \textbf{TAS} & \textbf{SH} & \textbf{DH}\\
    \midrule
    MoCo & 32.8\% & 21.3\% & 63.4\% & 2.9 y & 6.1 y \\
    SI & 30.8\% & 26.8\% & 80.1\% & 4.8 y & 6.0 y \\
    CORAL & 30.9\% & 24.8\% & 74.2\% & 4.9 y & 6.5 y \\
    ERM & 31.9\% & 25.7\% & 74.6\% & 4.5 y & 6.7 y \\
    A-GEM & 16.2\% & 6.9\% & 42.9\% & 1.7 y & 6.3 y \\
    FT & 32.5\% & 27.1\% & 75.9\% & 5.1 y & 6.4 y \\
    \bottomrule
  \end{tabular}
\end{table}

Table~\ref{table1} compares the six evaluated models, MoCo (with SML), Synaptic Intelligence (SI), CORAL, Empirical Risk Minimization (ERM), A-GEM, and Fine-Tuning (FT), across all proposed metrics. 
Traditional ID and OOD accuracies show overall consistency with the Wild-Time benchmark, while TAS, SH, and DH provide deeper insights into temporal adaptation. 
For instance, although MoCo+SML achieves a higher ID accuracy, its TAS and SH values are significantly lower, revealing poor retention of relative performance over time. 
In contrast, methods such as SI and FT exhibit stronger temporal robustness, maintaining higher TAS and SH despite similar absolute accuracy levels. 
Interestingly, the Drift Horizon remains consistently around six years for most models, indicating a common temporal limit beyond which adaptation becomes ineffective.

\subsection{Results on Yearbook}

\begin{table}
  \caption{Comparison of models on the Yearbook dataset (1930–2013). 
While ID and OOD averages indicate overall robustness, TAS highlights differences in temporal adaptability, particularly under distributional changes.}
  \label{table2}
  \begin{tabular}{l|l|l|l|l|l}
    \toprule
    \textbf{\makecell{Model}} & \textbf{\makecell{ID \\Avg}} & \textbf{\makecell{OOD \\Avg}} & 
    \textbf{\makecell{OOD \\Min}} & \textbf{\makecell{TAS \\Avg}} & 
    \textbf{\makecell{TAS \\Min}} \\
    \midrule
    MoCo & 95.0\% & 92.8\% & 76.8\% & 97.8\% & 89.3\% \\
    SI & 93.8\% & 90.9\% & 73.3\% & 96.2\% & 75.8\% \\
    CORAL & 91.0\% & 87.5\% & 61.3\% & 96.3\% & 82.5\% \\
    EWC & 95.2\% & 92.6\% & 70.3\% & 96.4\% & 72.9\% \\
    ERM & 94.7\% & 91.8\% & 70.3\% & 96.8\% & 73.1\% \\
    A-GEM & 90.4\% & 87.2\% & 56.5\% & 96.2\% & 61.0\% \\
    FT & 95.4\% & 92.4\% & 70.3\% & 96.2\% & 72.9\% \\
    \bottomrule
  \end{tabular}
\end{table}

\begin{figure*}[b]
    \centering
    \includegraphics[width=\textwidth]{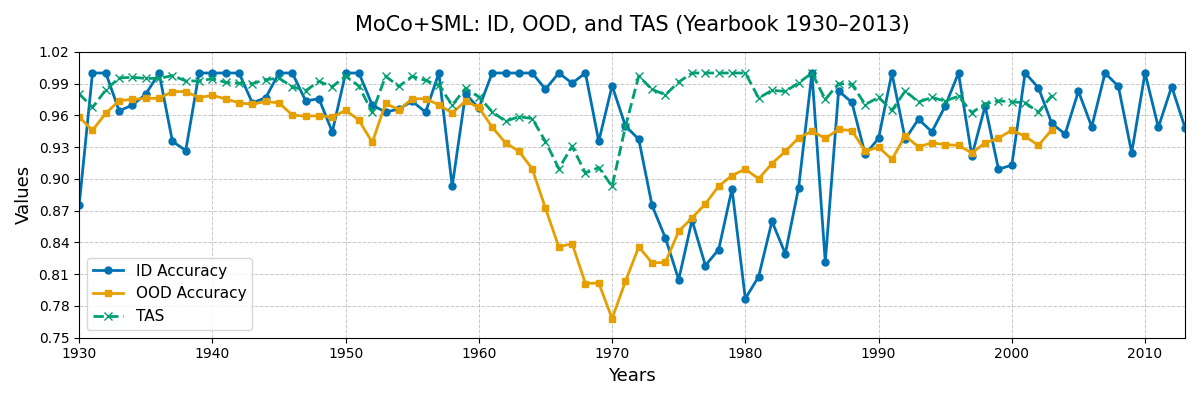}
    \caption{
    ID, OOD, and TAS trends for the MoCo+SML model on Yearbook (1930–2013). 
    Around 1970, a large apparent ID–OOD gap (99\% vs. 77\%) corresponds to a TAS of about 90\%, indicating that the performance drop is largely due to intrinsic data difficulty rather than failure to adapt.
    }
    \label{fig:yearbook}
\end{figure*}

The Yearbook dataset provides a long temporal span, making it a suitable benchmark to analyze gradual visual and distributional changes over decades. 
Table~\ref{table2} reports the average and worst-case results for the evaluated models across standard metrics (ID, OOD) and the proposed ones (TAS).
As observed in prior work, all models achieve strong ID and OOD accuracy, yet the proposed TAS metric provides additional insight into how well each model adapts to the temporal evolution of the data.

To better understand this distinction, Figure~\ref{fig:yearbook} analyzes the temporal behavior of \textit{MoCo+SML}. 
The plot shows the yearly evolution of In-Distribution (ID) accuracy, Out-of-Distribution (OOD) accuracy, and the corresponding Temporal Adaptation Score (TAS). 
A clear example of how TAS complements ID–OOD analysis appears around 1970: while ID and OOD accuracies exhibit a sharp gap, approximately from 99\% to 77\%, the TAS value at that point remains close to 90\%. 
This indicates that, although the apparent drop suggests poor generalization, the model still retains 90\% of its oracle performance. 
In other words, the degradation is not purely due to lack of temporal adaptation, but rather to an intrinsic increase in data difficulty, as also reflected by the decline in ID accuracy observed in subsequent years. 
TAS thus allows distinguishing between a genuine adaptation failure and a natural evolution of the task itself, revealing that models may remain relatively well-aligned with the underlying temporal dynamics even when absolute performance decreases.

\section{Conclusions and Future Work}

This work introduced a framework for evaluating temporal adaptation in machine learning models. 
We argued that existing metrics—such as In-Distribution (ID) and Out-of-Distribution (OOD) accuracy—confuse intrinsic data difficulty with a model’s actual capacity to adapt over time. 
To address this ambiguity, we proposed three complementary post-hoc metrics: the Temporal Adaptation Score (TAS), the Stability Horizon (SH), and the Drift Horizon (DH). 
Together, they provide a dynamic and interpretable view of model robustness under temporal distribution shift.

Experiments on the Yearbook and FMoW benchmarks demonstrated that these metrics uncover adaptation patterns that remain hidden under static ID–OOD comparisons. 
In particular, TAS captures relative adaptation rather than absolute accuracy, distinguishing genuine temporal misalignment from intrinsic degradation in data quality. 
The Stability Horizon quantifies how long a model remains reliable, while the Drift Horizon identifies when degradation becomes statistically significant. 
Our results suggest that what has often been interpreted as a persistent failure to generalize over time may, in many cases, reflect a slower but still effective adaptation process.

Several extensions are worth pursuing. 
First, while our metrics are post-hoc and require full temporal supervision, future research could investigate online approximations that estimate adaptation dynamics during deployment. 
Second, integrating these metrics with temporal model selection or active retraining strategies could enable automatic detection of retraining points, reducing computational costs while maintaining accuracy. 
Third, exploring their applicability beyond classification (e.g., to regression, forecasting, or multimodal temporal learning) would help assess their generality across domains. 
Finally, a deeper theoretical analysis of the relationship between temporal shift magnitude, adaptation speed, and stability horizons could lead to more formal guarantees of temporal robustness.

Overall, our findings highlight that evaluating adaptation under temporal distribution shifts requires going beyond static accuracy metrics. 
By explicitly modeling how performance evolves over time, we can move toward a more faithful understanding of model behavior in truly dynamic environments.

%% The declaration on generative AI comes in effect
%% https://ceur-ws.org/GenAI/Policy.html
\section*{Declaration on Generative AI}
 During the preparation of this work, the authors used ChatGPT (by OpenAI) for grammar and spelling checks. After using this tool, the authors reviewed and edited the content as needed and take full responsibility for the publication’s content.

%%
%% Define the bibliography file to be used
\bibliography{sample-ceur}

\end{document}